\documentclass[10pt,twocolumn,letterpaper]{article}

\usepackage[pagenumbers]{cvpr} 

\definecolor{cvprblue}{rgb}{0.21,0.49,0.74}
\usepackage[pagebackref,breaklinks,colorlinks,allcolors=cvprblue]{hyperref}
\usepackage{multirow}
\usepackage[accsupp]{axessibility}  

\def\paperID{} 
\def\confName{CVPR}
\def\confYear{2026}

\title{Temporally Consistent Long-Term Memory for 3D Single Object Tracking}

\author{Jaejoon Yoo, SuBeen Lee, Yerim Jeon, Miso Lee, Jae-Pil Heo\thanks{Corresponding author}\\
Sungkyunkwan University\\
{\tt\small \{dbwowns10, leesb7426, 1357j, dlalth557, jaepilheo\}@skku.edu}\\
}

\begin{document}
\maketitle
\begin{abstract}
3D Single Object Tracking (3D-SOT) aims to localize a target object across a sequence of LiDAR point clouds, given its 3D bounding box in the first frame.
Recent methods have adopted a memory-based approach to utilize previously observed features of the target object, but remain limited to only a few recent frames.
This work reveals that their temporal capacity is fundamentally constrained to short-term context due to severe temporal feature inconsistency and excessive memory overhead.
To this end, we propose a robust long-term 3D-SOT framework, \textit{ChronoTrack}, which preserves the temporal feature consistency while efficiently aggregating the diverse target features via long-term memory.
Based on a compact set of learnable memory tokens, \textit{ChronoTrack} leverages long-term information through two complementary objectives: a temporal consistency loss and a memory cycle consistency loss.
The former enforces feature alignment across frames, alleviating temporal drift and improving the reliability of proposed long-term memory.
In parallel, the latter encourages each token to encode diverse and discriminative target representations observed throughout the sequence via memory-point-memory cyclic walks.
As a result, \textit{ChronoTrack} achieves new state-of-the-art performance on multiple 3D-SOT benchmarks, demonstrating its effectiveness in long-term target modeling with compact memory while running at real-time speed of 42 FPS on a single RTX 4090 GPU.
The code is available at \url{https://github.com/ujaejoon/ChronoTrack}
\end{abstract}
\vspace{-0.3cm}
\section{Introduction}
3D Single Object Tracking (3D-SOT) is a fundamental task in computer vision, essential for various applications such as robotics and autonomous driving ~\cite{robotics1, robotics2, autodrive1}.
Given a target object annotated with a 3D bounding box in the first frame, 3D-SOT aims to predict the corresponding object's bounding box in subsequent frames of point clouds.

\begin{figure}[!t]
    {\includegraphics[width=0.49\textwidth]{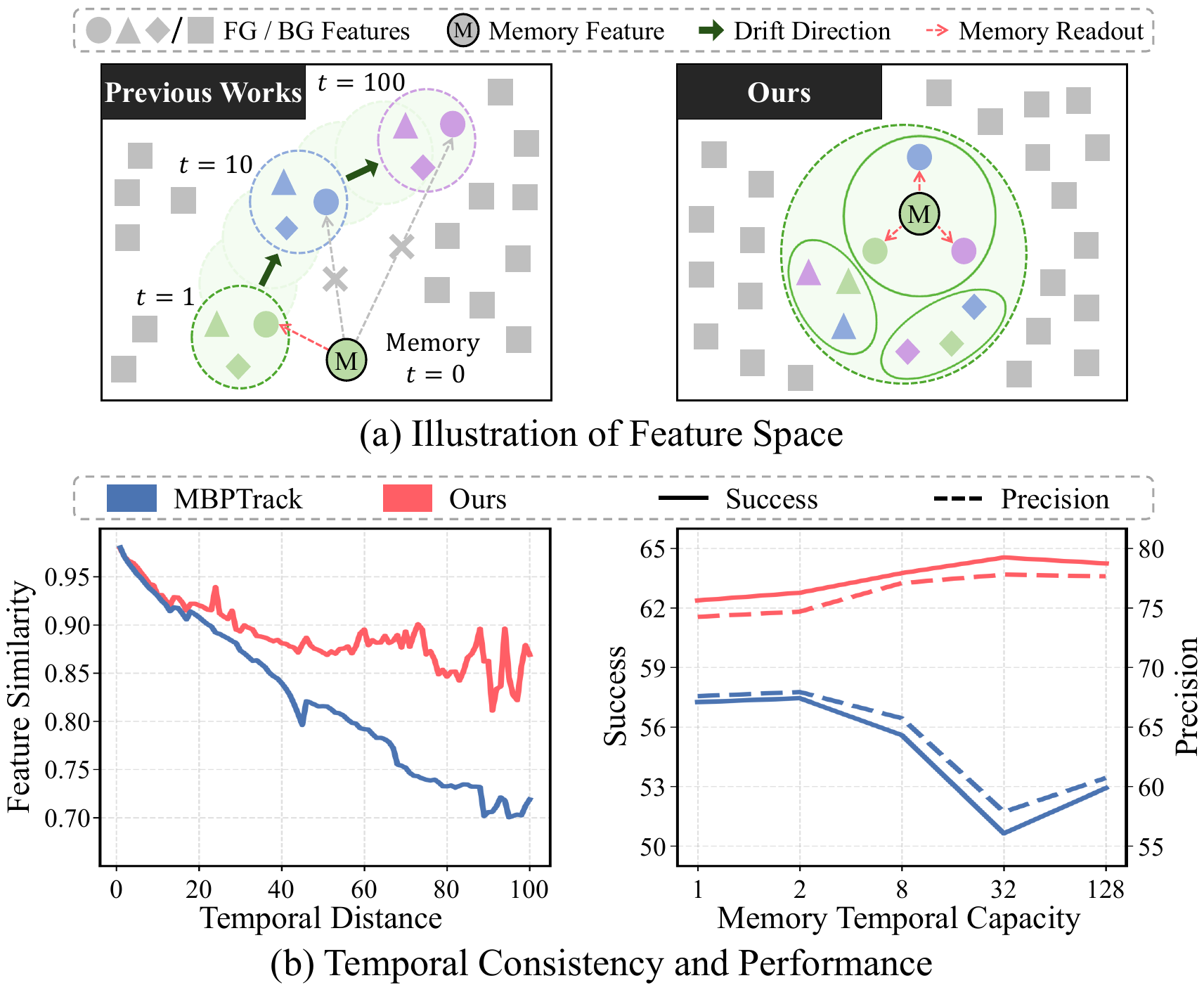}}
    \centering
    \caption{
    \textbf{(a)} Illustration of the feature space in existing works versus ours.
    Different shapes and colors represent different parts of the object and different time indices, respectively.
    In existing methods, target features tend to drift as the target's appearance changes, resulting in temporal inconsistency that diminishes the utility of earlier features in memory.
    In contrast, our method enforces temporal feature consistency across frames, mitigating drift and enabling more reliable memory readout.
    \textbf{(b)} Quantitative analysis of temporal feature consistency and its correlation with performance.
    We evaluate the temporal consistency by measuring the average cosine similarity between target features from each frame and those from a future frame.
    Compared to MBPTrack, our method maintains higher feature similarity as the temporal distance increases.
    This robustness allows our method to take advantage of longer context, whereas MBPTrack fails to benefit from increased temporal capacity and instead exhibits performance (Success and Precision) degradation.
    }
\label{fig:fig_1}
\vspace{-0.4cm}
\end{figure}
Recent methods ~\cite{MBPTrack, M3SOT, StreamTrack, HVTrack} adopt a memory-based approach, storing point-level representations of a few recent frames to propagate the target cue into future frames.
By leveraging temporal context from multiple historical frames, they have made significant advances in the field.
Despite this progress, they still rely on a short-term context of only two or three recent frames, which motivates us to exploit longer temporal context.

One straightforward approach to capture long-term context is simply extending the memory length.
However, such na\"ive extension presents two fundamental challenges.
First, it does not consistently improve tracking performance, as illustrated in Fig.~\ref{fig:fig_1}.
We observe that feature similarity between corresponding target points across frames declines as the temporal gap increases, and incorporating such low-similarity frames into memory leads to degraded performance.
This issue has been underexplored, as existing methods~\cite{MBPTrack, MemDisst, StreamTrack} typically have been confined to short temporal windows, where the appearance variations of the target are relatively minor.
Furthermore, extending point-level memory leads to a substantial increase in memory cost, as more point features must be stored and processed for prediction, as shown in Fig.~\ref{fig:fig_7}.
This cost grows proportionally with the length of memory, making such approaches impractical for real-world applications on edge devices.

To address these issues, we propose \textit{ChronoTrack}, a robust long-term 3D-SOT framework that captures diverse aspects of the target while effectively leveraging long-range temporal cues.
Specifically, we construct a compact set of learnable memory tokens that are recurrently updated at each timestep by integrating new observations of the target with previously stored contexts.
Moreover, to improve the utility of long-range information captured by our token-level memory, we introduce two complementary objectives: temporal consistency loss and memory cycle consistency loss.
Temporal consistency loss enforces feature alignment across time, mitigating feature drift caused by appearance variations and encouraging the effective use of long-term context.
Building on these temporally aligned representations, we further aim to encode the diverse appearances of the target observed over time into the memory. 
To this end, we encourage consistent associations between memory tokens and semantically distinct target parts through a cyclic walk mechanism.
Consequently, our token-level memory provides compact yet expressive representations that consistently capture reliable and diverse target cues without incurring excessive overhead.


In summary, our contributions are as follows:
\begin{itemize}
    \item We propose \textit{ChronoTrack}, a long-term 3D-SOT framework where a compact memory module recurrently integrates recent observations with accumulated target context.
    \item We design a temporal consistency loss that enforces temporal feature consistency, thereby improving the reliability of long-term memory.
    \item We introduce a memory cycle consistency loss that promotes diversity among tokens by encouraging each one to represent semantically distinct parts of the object.
    \item Our approach outperforms previous memory-based methods on standard 3D-SOT benchmarks, validating its effectiveness in taking advantage of the long-term context.
\end{itemize}
\section{Related Work}
A central challenge in 3D Single Object Tracking (3D-SOT) is how to effectively leverage past observations to propagate the target cue into future frames.

Early approaches~\cite{SC3D, P2B, V2B, MLVSNet, BAT, PTTR, STNet} adopted a template-based strategy, where the target is represented by a cropped point cloud from a single frame.
They compute point-wise feature similarity with the template and subsequently predict multiple bounding box proposals using a 3D region proposal network. 
Follow-up works extended this framework, such as incorporating background information~\cite{CXTrack} and predicting relative motion between consecutive frames~\cite{M2Track}.
However, these methods still rely on a single reference frame.
Therefore, when the target is occluded in that frame, these approaches struggle to maintain reliable tracking due to the lack of sufficient target cues.

To overcome these limitations, memory-based methods such as MBPTrack~\cite{MBPTrack}, HVTrack~\cite{HVTrack}, M3SOT~\cite{M3SOT}, StreamTrack~\cite{StreamTrack}, and MemDisst~\cite{MemDisst} adopt an external memory to save information from historical frames.
They store point-level representations from multiple past frames to localize the target object in the current frame, improving robustness to occlusion and appearance changes.
However, they are still confined to short-term memory and often fail to benefit from longer temporal context as memory length increases.
We find that this limitation primarily arises from temporal feature inconsistency, where target representations become misaligned across distant frames, reducing the effectiveness of stored contexts.
To address this issue, our method promotes temporal consistency across frames, enabling more reliable use of information accumulated over longer durations.
\section{Method}
\begin{figure*}[t]
    \centering
    {\includegraphics[width=0.98\textwidth]{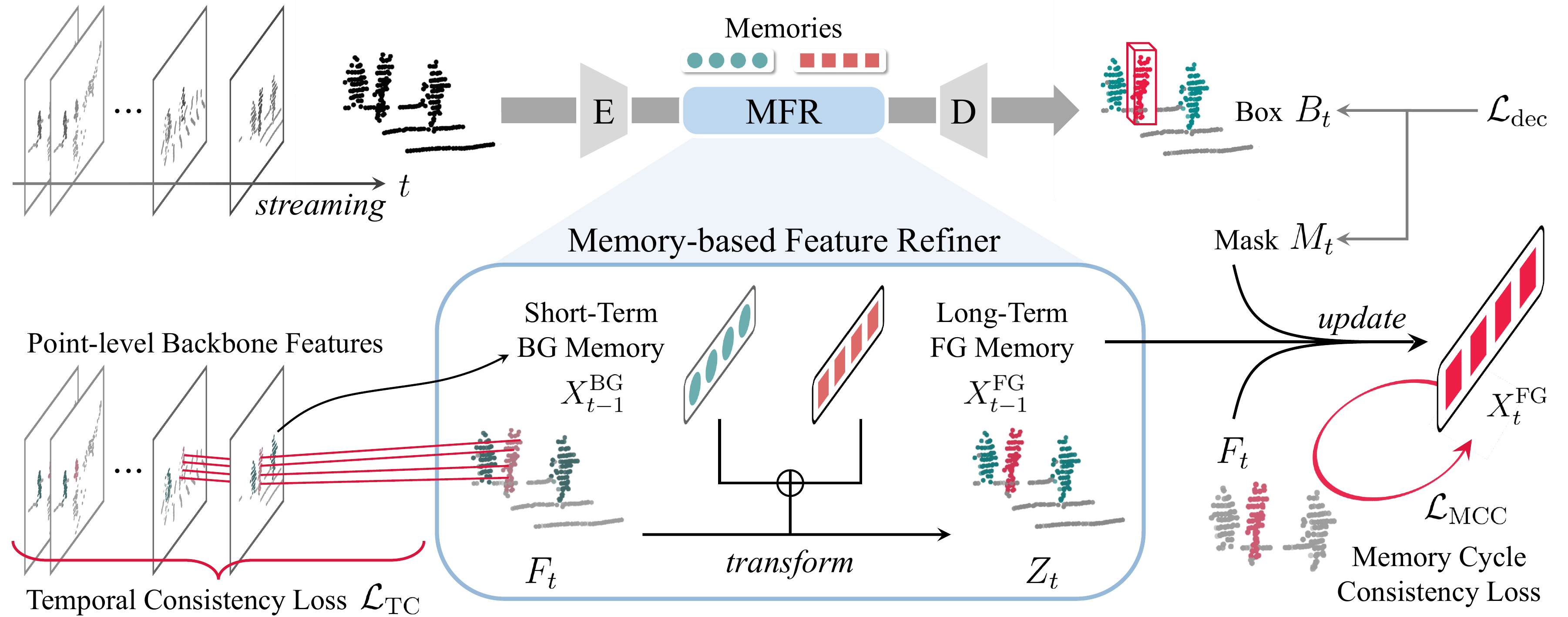}}
    \caption{
    \textbf{Overall pipeline of \textit{ChronoTrack}.}
    The point features $F_t$ of the current point cloud $P_t$ are first extracted by the backbone network $E$.
    These features are then fed to the Memory-based Feature Refiner~(MFR) along with the long-term target memory $X^{\text{FG}}_{t-1}$ and short-term background memory $X^{\text{BG}}_{t-1}$, injecting long-term target cues and short-term contextual information into the current features, resulting in target-aware features $Z_t$.
    The decoder~$D$ then predicts the 3D bounding box $B_t$ and targetness mask $M_t$ from these enriched features, and the memory is updated based on the predicted mask and current point features.
    To ensure the reliability of accumulated information in memory and foster memory diversity, two complementary losses are applied: temporal consistency loss $\mathcal{L}_{\text{TC}}$ and memory cycle consistency loss $\mathcal{L}_{\text{MCC}}$.
    }
\label{fig:fig_2}
\end{figure*}
\subsection{Overview}

\paragraph{Task Definition.}
Given a target object annotated with the 3D bounding box in the first frame, 3D Single Object Tracking~(3D-SOT) aims to estimate the bounding boxes of the target in subsequent frames.
At timestep $t \in [2, T]$, the frame consists of a point cloud $P_t \in \mathbb{R}^{N_t \times 3}$ within the search area, where $N_t$ is the number of points, and a bounding box $B_t \in \mathbb{R}^7$ is defined by its center $(x_t, y_t, z_t)$, heading angle around up-axis $\theta_t$, and size $(w, l, h)$, which is fixed across frames.

\paragraph{Overall Pipeline.}
Our overall pipeline is illustrated in Fig.~\ref{fig:fig_2}.
Unlike existing memory-based methods that rely only on short-term context, our approach captures long-term temporal cues of the target (i.e. foreground) through a compact set of learnable memory tokens.
At each timestep, current point features are transformed into target-aware features by attending to long-term foreground memory and short-term background memory, and fed into the decoder for prediction.
After prediction, foreground memory tokens are updated by aggregating features of predicted target points, thereby maintaining both recent and past context.
To enhance the efficacy of this long-term context within tokens, we first enforce temporal feature consistency of the target features.
Specifically, we introduce a temporal consistency loss to promote high similarity between corresponding parts of target points across frames, mitigating temporal drift even under severe appearance changes.
Furthermore, as the target is naturally observed under diverse appearances throughout long-term, we seek to encode this variability into memory tokens.
To achieve this, we propose a memory cycle consistency loss that reinforces the association between each token and distinct parts of the target by conducting a two-step cyclic walk between memory tokens and point features.
\begin{figure*}[t]
    \centering
    {\includegraphics[width=0.98\textwidth]{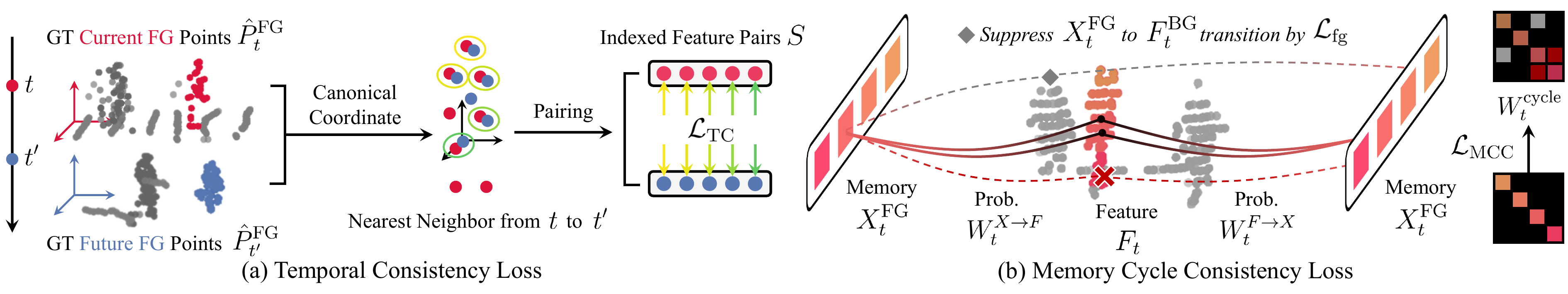}}
    \caption{
    \textbf{Losses for temporal consistency and token diversity.}
    (a) Target points from distant frames are transformed into canonical coordinates to identify approximate pairs of spatially aligned points that likely belong to the same part of the target object.
    A temporal consistency loss is then applied to each pair to enforce high feature similarity.
    (b) On the basis of these temporally aligned features, each memory token undergoes a two-step walk (memory $\rightarrow$ points $\rightarrow$ memory).
    The walk is optimized to simultaneously maximize the probability that each token returns to itself after a two-step walk and the likelihood of transiting through foreground points, thereby encouraging tokens to encode diverse and target-specific semantics.
    }
    \vspace{-0.3cm}
\label{fig:fig_3}
\end{figure*}
\subsection{Long-Term Memory}

Given a point cloud $P_t \in \mathbb{R}^{N_t \times 3}$ at time step $t$, we extract point-wise features $F_t \in \mathbb{R}^{N_t \times D}$ where $N_t$ and $D$ represent the number of points and the feature dimension, respectively.
For simplicity, we omit the notation of point downsampling.

Unlike previous works that rely solely on short-term temporal context, we leverage the long-term context of the target object, spanning from the initial frame to the most recent one.
To this end, we define a compact set of $K$ learnable memory tokens $X^\text{FG}_0 \in \mathbb{R}^{K \times D}$.
To initialize these tokens, we extract ground-truth foreground point features of the first frame $\hat{F}^{\text{FG}}_1 \in \mathbb{R}^{\hat{N}^{\text{FG}}_1 \times D}$ at the beginning of each tracking sequence, as follows: 
\begin{equation}
    \begin{split}
        \label{eq:foreground_points_gt}
        \hat{F}_1^{\text{FG}} = \{F_{1,i} \mid \hat{M}_{1,i}=1\}, \quad \forall i=1, \cdots\!, N_1,
    \end{split}
\end{equation}
where $\hat{M}_t \in \{0,1\}^{N_t}$ is the ground-truth mask of $t$-th frame, that $1$ indicates target points and $0$ denotes background.
Then, we compute initial tokens $X^\text{FG}_1 \in \mathbb{R}^{K \times D}$, as follows:
\begin{equation}
    \begin{split}
        \label{eq:token_initialize}
        X^\text{FG}_{1} &= \text{MU}(X^\text{FG}_{0}, \hat{F}_1^{\text{FG}}, \hat{F}_1^{\text{FG}}),
    \end{split}
\end{equation}
where $\text{MU}$ denotes the memory updater with $L_{\text{MU}}$ layers, each comprising cross-attention, self-attention, and MLP. $(\cdot,\cdot,\cdot)$ correspond to \textit{query}, \textit{key}, and \textit{value} of cross-attention.

Once initialized, foreground memory tokens are recurrently updated to accumulate target features, aggregating the long-term context of the target. 
However, these tokens focus exclusively on the target object, while background contexts are also crucial for accurate target localization~\cite{CXTrack, MBPTrack}.
To complement this, we introduce a short-term background memory that stores background point features from the immediately preceding frame.
The initial background memory $X^{\text{BG}}_1 $ is defined with the ground-truth background point features of the first frame $\hat{F}^{\text{BG}}_1 \in \mathbb{R}^{\hat{N}^\text{BG}_1 \times D}$, as follows:
\begin{equation}
    \begin{split}
        \label{eq:background_points_pred}
        X^\text{BG}_1 = \{F_{1,i} \mid \hat{M}_{1,i} = 0\}, \quad \forall i = 1,\cdots\!,N_1.
    \end{split}
\end{equation}
By utilizing both foreground memory $X^\text{FG}_{t-1}$ and background memory $X^\text{BG}_{t-1}$, the current point features $F_t$ are transformed into target-aware point features $Z_t \in \mathbb{R}^{N_{t} \times D}$, as follows:
\begin{equation}
    \label{eq:feature_propagation}
    Z_{t} = \text{MFR}(F_{t}, [X^\text{FG}_{t-1}, X^\text{BG}_{t-1}], [X^\text{FG}_{t-1}, X^\text{BG}_{t-1}]),
\end{equation}
where $\text{MFR}$ is the memory-based feature refiner, comprising $L_{\text{MFR}}$ layers of cross-attention, self-attention, and an MLP.
Here, $(\cdot,\cdot,\cdot)$ denotes \textit{query}, \textit{key}, and \textit{value} of cross-attention, and $[\cdot,\cdot]$ is concatenation along sequence dimension.

Next, the decoder receives these transformed features and predicts the point-wise targetness score $M_{t} \in \mathbb{R}^{N_t}$ and the 3D bounding box $B_{t} \in \mathbb{R}^7$.
After prediction, the foreground memory tokens are updated to integrate both themselves that encode past contexts and the features of predicted target points $F_{t}^{\text{FG}} \in \mathbb{R}^{N^\text{FG}_t \times D}$ in the current frame as follows:
\begin{equation}
    \begin{split}
        \label{eq:foreground_points_pred}
        F_{t}^{\text{FG}} = \{F_{t,i} \mid M_{t,i} \geq \tau_{\text{mask}}\}, \: \forall i = 1,\cdots\!,N_t,
    \end{split}
\end{equation}
\vspace{-0.4cm}
\begin{equation}
    \begin{split}
        \label{eq:token_update}
        X^\text{FG}_{t} = \text{MU}(X^\text{FG}_{t-1}, [X^\text{FG}_{t-1},F^\text{FG}_{t}], [X^\text{FG}_{t-1},F^\text{FG}_{t}]),
    \end{split}
\end{equation}
where $\tau_{\text{mask}}$ and $X^\text{FG}_t \in \mathbb{R}^{K \times D}$ denote the mask threshold and the updated foreground memory tokens, respectively.
On the other hand, the background memory is substituted by the newly predicted background features, as follows:
\begin{equation}
    \begin{split}
        \label{eq:background_update}
        X_{t}^{\text{BG}} = \{F_{t,i} \mid M_{t,i} < \tau_{\text{mask}}\}, \: \forall i = 1,\cdots\!,N_t,
    \end{split}
\end{equation}
where $X^\text{BG}_t \in \mathbb{R}^{N^\text{BG}_t \times D}$ is the updated background memory.

The key advantage of our token-based memory design is the adaptive integration of newly observed and temporally accumulated target features into fixed-sized memory tokens.
This allows ChronoTrack to leverage long-term target context even under a limited memory budget, unlike existing point-level memory methods~\cite{MBPTrack, HVTrack, StreamTrack, M3SOT, MemDisst}.
\subsection{Temporal Consistency Loss}
To effectively utilize the long-term temporal context of our token-level memory, the feature representations of the target object should be consistent over time, as illustrated in Fig.~\ref{fig:fig_1}.
To achieve this, we propose a temporal consistency loss that explicitly encourages part-level consistency by aligning features of spatially corresponding target points across time.
This enables the model to retain structurally consistent features despite changes in viewpoint and partial occlusions.

Specifically, we first transform ground-truth foreground points $\hat{P}_t^\text{FG} \in \mathbb{R}^{\hat{N}_t^\text{FG} \times 3}$ into a canonical coordinate system $C^\text{FG}_t \in \mathbb{R}^{\hat{N}_t^\text{FG} \times 3}$, as follows:
\begin{equation}
    \begin{split}
        \label{eq:canonical_transform}
        C^\text{FG}_t = (\hat{P}_t^\text{FG} - \hat{\mathbf{c}}_t)\hat{R}_t,
    \end{split}
\end{equation}
where $\hat{R}_t \in \mathbb{R}^{3 \times 3}$ and $\hat{\mathbf{c}}_t \in \mathbb{R}^{3}$ are the rotation matrix derived from the heading angle and the center of the ground-truth bounding box, respectively, and the subtraction by $\hat{\mathbf{c}}_t$ is applied row-wise.

Based on the canonical coordinates, we establish point-level correspondences by computing the nearest neighbors in each future frame, as follows:
\begin{equation}
    \begin{split}
        \label{eq:point_correspondence}
        j_{t,i,t'} = \underset{1 \leq k \leq \hat{N}^{\text{FG}}_{t'}}{\arg\min} \| C^{\text{FG}}_{t,i} - C^{\text{FG}}_{t',k} \|_2, \; \forall t' = t+1, \dots, T.
    \end{split}
\end{equation}
For each future frame $t'$ from $t{+}1$ to $T$ in the training sequence, we construct a set of indices with reliable correspondences by including only the pairs whose canonical distance is below the threshold $\tau_{\text{dist}}$, thereby filtering out unreliable matches.
\begin{equation}
    \begin{split}
        S =
    \left\{ (t,i,t') \mid \| C^\text{FG}_{t,i} - C^{\text{FG}}_{t',j_{t,i,t'}} \|_2 < \tau_{\text{dist}} \right\}.
    \end{split}
\end{equation}
By matching points that occupy similar positions in the canonical system, we obtain corresponding point pairs, which represent approximately the same part of the target object across the different frames.
As a result, we define the temporal consistency loss to encourage feature consistency over time, as follows:
\begin{equation}
    \begin{split}
        \label{eq:temporal_consistency_loss}
        \mathcal{L}_\text{TC} = \frac{1}{|S|}\sum_{(t,i,t') \in S} \text{SmoothL1} (\hat{F}_{t,i}^\text{FG}, \hat{F}_{t',j_{t,i,t'}}^\text{FG}),
    \end{split}
\end{equation}
where $\text{SmoothL1}$ is smooth L1 loss function~\cite{SmoothL1}.

By enforcing feature alignment across frames, we ensure that features from earlier frames encoded in our token-level memory remain semantically consistent over time, as depicted in Fig.~\ref{fig:fig_1}.
This is particularly crucial in scenarios such as occlusions or viewpoint shifts, where drastic appearance changes cause feature drift for the same object parts.
\subsection{Memory Cycle Consistency Loss}
Building upon temporally aligned features, we further enhance our memory representation through a memory cycle consistency loss that promotes token-level diversity.
While long-term tracking naturally exposes the target under varying views, we aim to render these varied appearances into memory tokens that capture diverse semantics of the target.

To this end, we draw inspiration from cyclic walk formulation in self-supervised learning~\cite{cyclicslot, randomwalk}, and enforce cycle consistency between memory tokens and point-level features to guide the memory update process.  
This objective encourages each token to establish stable and localized associations with distinct semantic parts of the target.

Specifically, we compute the transition probability matrix from tokens to points $W^{X \rightarrow F}_t \in \mathbb{R}^{K \times N_t}$ and from points back to tokens $W^{F \rightarrow X}_t \in \mathbb{R}^{N_t \times K}$, as follows:
\begin{equation}
    \begin{split}
        \label{eq:transition_1}
        W^{X \rightarrow F}_{t,i,j} =
        \frac{\exp(\text{sim}(X^\text{FG}_{t,i} \cdot F_{t,j}) / \tau_\text{cycle})}{\sum_{k=1}^{N_t} \exp(\text{sim}(X^\text{FG}_{t,i} \cdot F_{t,k}) / \tau_\text{cycle})},
    \end{split}
\end{equation}
\vspace{-0.2cm}
\begin{equation}
    \begin{split}
        \label{eq:transition_2}
        W^{F \rightarrow X}_{t,i,j} =
        \frac{\exp( \text{sim} (F_{t,i} \cdot X^\text{FG}_{t,j}) / \tau_\text{cycle})}{\sum_{k=1}^{K} \exp( \text{sim} (F_{t,i} \cdot X^\text{FG}_{t,k}) / \tau_\text{cycle})},
    \end{split}
\end{equation}
where $\text{sim}(\cdot)$ refers to the cosine similarity between two vectors and $ \tau_\text{cycle}$ is the temperature.
Then, we derive a two-step cyclic transition matrix $W^\text{cycle}_t \in \mathbb{R}^{K \times K}$, where each diagonal entry indicates the probability of a memory token returning to itself through the cycle, as follows:
\begin{equation}
    \begin{split}
        \label{eq:cyclic_transition}
        W^\text{cycle}_t = W^{X \rightarrow F}_{t}W^{F \rightarrow X}_{t}.
    \end{split}
\end{equation}

Based on this two-step cyclic walk, we aim to guide each memory token to encode non-redundant and discriminative features of the target.
To achieve this, we propose a memory cycle consistency loss $\mathcal{L}_{\text{MCC}}$, which is composed of the following two terms:
\begin{equation}
    \begin{split}
        \label{eq:cycle_consistency_loss}
        \mathcal{L}_\text{cycle} = \frac{1}{T}\sum^T_{t=1} \text{CE}(W^\text{cycle}_t, I),
    \end{split}
\end{equation}
\begin{equation}
\label{eq:loss_fg}
\mathcal{L}_{\text{fg}} = -\frac{1}{T} \frac{1}{K} \sum_{t=1}^{T} \sum_{i=1}^{K} \log \Big( \!\! \sum_{j \in \hat{\mathcal{I}}_t^{\text{FG}}} W_{t,i,j}^{X \rightarrow F} \Big),
\end{equation}
\begin{equation}
\label{eq:memory_cycle_consistency_loss}
\mathcal{L}_{\text{MCC}} = \mathcal{L}_{\text{cycle}} + \mathcal{L}_{\text{fg}},
\end{equation}
where $\text{CE}(\cdot,\cdot)$ denotes the cross-entropy loss, $I \in \mathbb{R}^{K\times K}$ is the identity matrix,  
and $\hat{\mathcal{I}}_t^{\text{FG}}$ denotes the set of ground-truth foreground point indices at time $t$.
Intuitively, $\mathcal{L}_{\text{cycle}}$ facilitates each token returning to itself through a two-step cycle by enforcing that the cyclic walk $W_t^{\text{cycle}}$ should be an identity matrix $I$.
To enable such self-return, each token must encode a semantically distinct part of the target.
Concurrently, $\mathcal{L}_{\text{fg}}$ encourages the transition path to pass through the foreground regions during the cycle.
By jointly minimizing both terms, memory tokens are optimized to represent a certain part of the target while emphasizing target-specific semantics over the background, which is crucial for maintaining robust long-term target representations.
\begingroup
\setlength{\tabcolsep}{9pt} 
\renewcommand{\arraystretch}{1} 
\begin{table*}[!t]
    \centering
    \caption{Comparison with the SOTA methods on KITTI. ``Mean" denotes the average results weighted by frame numbers. \textbf{Bold} and \underline{underline} represent the best and second-best results, respectively.}
    \label{tab:kitti}
    {
    \begin{tabular}{l|cc|cc|cc|cc|cc}
        \hline
        \multirow{2}{*}{Method} &
        \multicolumn{2}{c|}{Car} &
        \multicolumn{2}{c|}{Pedestrian} &
        \multicolumn{2}{c|}{Van} &
        \multicolumn{2}{c|}{Cyclist} &
        \multicolumn{2}{c}{Mean} \\
        \cline{2-11}
        & Suc. & Prec.
        & Suc. & Prec.
        & Suc. & Prec.
        & Suc. & Prec.
        & Suc. & Prec. \\
        \hline
        SC3D~\cite{SC3D} & 41.3 & 57.9 & 18.2 & 37.8 & 40.4 & 47.0 & 41.5 & 70.4 & 31.2 & 48.5 \\
        P2B~\cite{P2B} &56.2 & 72.8 & 28.7 & 49.6 & 40.8 & 48.4 & 32.1 & 44.7 & 42.4 & 60.0 \\
        STNet~\cite{STNet} & 72.1 & 84.0 & 49.9 & 77.2 & 58.0 & 70.6 & 73.5 & 93.7 & 61.3 & 80.1 \\
        M$^2$-Track~\cite{M2Track} & 65.5 & 80.8 & 61.5 & 88.2 & 53.8 & 70.7 & 73.2 & 93.5 & 62.9 & 83.4 \\ 
        CXTrack~\cite{CXTrack} & 69.1 & 81.6 & 67.0 & 91.5 & 60.0 & 71.8 & 74.2 & 94.3 & 67.5 & 85.3 \\
        MBPTrack~\cite{MBPTrack} & 73.4 & 84.8 & 68.6 & 93.9 & 61.3 & 72.7 & 76.7 & 94.3 & 70.3 & 87.9 \\
        SCVTrack~\cite{SCVTrack} &
        68.7 & 81.9 & 62.0 & 89.1 & 58.6 & 72.8 & 77.4 & 94.4 & 65.1 & 84.5 \\
        StreamTrack~\cite{StreamTrack} & 72.6& 83.7 & \textbf{70.5} & \textbf{94.7} & 61.0 & 76.9 & \textbf{78.1} & 94.6 & 70.8 & 88.1 \\
        M3SOT~\cite{M3SOT} & \underline{75.9} & \underline{87.4} & 66.6 & 92.5 & 59.4 & 74.7 & 70.3 & 93.4 &  70.3 & 88.6 \\
        HVTrack~\cite{HVTrack} & 68.2 & 79.2 & 64.6 & 90.6 & 54.8 & 63.8 & 72.4 & 93.7 & 65.5 & 83.1 \\
        MemDisst~\cite{MemDisst} & 74.1 & 85.6 & \underline{69.1} & \underline{94.1} & \textbf{66.6} & \textbf{79.3} & 77.2 & \underline{94.7} & \underline{71.3} & \underline{88.9} \\  
        \hline
        ChronoTrack 
        & \textbf{76.0} & \textbf{88.6} 
        & 68.6 & 94.0 
        & \underline{64.2} & \underline{77.7} 
        & \underline{77.7} & \textbf{94.8} 
        & \textbf{71.8} & \textbf{90.1} \\
        \hline
    \end{tabular}
    
    }
\end{table*}
\endgroup 
\subsection{Training Objective}
We adopt the BPLocNet decoder from MBPTrack~\cite{MBPTrack} and apply its loss formulation without modification. 
To be specific, the decoder loss comprises five terms: a cross-entropy loss for the predicted point-wise targetness score $\mathcal{L}_{\text{m}}$, an MSE loss for the target center $\mathcal{L}_{\text{c}}$, cross-entropy losses for proposal quality and targetness scores of bbox proposals ($\mathcal{L}_{\text{q}}$, $\mathcal{L}_{\text{s}}$), and a smooth L1 loss for bounding box regression $\mathcal{L}_{\text{bbox}}$.
The overall decoder loss is given by:
\begin{equation}
\mathcal{L}_{\text{dec}} = 
\lambda_{\text{m}} \mathcal{L}_{\text{m}} + 
\lambda_{\text{c}} \mathcal{L}_{\text{c}} + 
\lambda_{\text{q}} \mathcal{L}_{\text{q}} + 
\lambda_{\text{s}} \mathcal{L}_{\text{s}} + 
\mathcal{L}_{\text{bbox}}.
\end{equation}
Our final training objective consists of the aforementioned decoder loss \(\mathcal{L}_{\text{dec}}\), temporal consistency loss \(\mathcal{L}_{\text{TC}}\) and memory cycle consistency loss \(\mathcal{L}_{\text{MCC}}\).
The overall loss is defined as:
\begin{equation}
\mathcal{L} = \mathcal{L}_{\text{dec}} + 
\mathcal{L}_{\text{TC}} + 
\mathcal{L}_{\text{MCC}}.
\end{equation}
\vspace{-0.3cm}
\section{Experiments}
\begingroup
\setlength{\tabcolsep}{6pt} 
\renewcommand{\arraystretch}{1.0} 
\begin{table*}[!t]
    \centering
    \caption{Comparison with the SOTA methods on NuScenes.}
    \label{tab:nuScenes}
    {
    \begin{tabular}{l|cc|cc|cc|cc|cc|cc}
        \hline
        \multirow{2}{*}{Method} &
        \multicolumn{2}{c|}{Car} &
        \multicolumn{2}{c|}{Pedestrian} &
        \multicolumn{2}{c|}{Truck} &
        \multicolumn{2}{c|}{Trailer} &
        \multicolumn{2}{c|}{Bus} &
        \multicolumn{2}{c}{Mean} \\
        \cline{2-13}
        & Suc. & Prec.
        & Suc. & Prec.
        & Suc. & Prec.
        & Suc. & Prec.
        & Suc. & Prec.
        & Suc. & Prec. \\
        \hline
        SC3D~\cite{SC3D} & 22.3 &  21.9 & 11.3 & 12.7 & 30.7 & 27.7 & 35.3 &  28.1 & 29.4 & 24.1  & 20.7 &  20.2 \\
        P2B~\cite{P2B} & 38.8 & 42.2 & 28.4 & 52.2 & 43.0 & 41.5 & 49.0 & 40.1 & 33.0 & 27.4 & 36.5 & 45.1 \\
        M$^2$-Track~\cite{M2Track}  & 55.9 & 65.1 & 32.1 & 60.9 & 57.4 & 59.5 & 57.6 & 58.3 & 51.4 & 51.4 & 49.2 & 62.7 \\
        MBPTrack~\cite{MBPTrack} & \underline{62.5} & 70.4 & \underline{45.3} & \underline{74.0} & 62.2 & 63.3 & 65.1 & 61.3 & 55.4 & 51.8 & \underline{57.5} & \underline{69.9}\\
        SCVTrack~\cite{SCVTrack} & 58.9 & 67.7 & 34.5 & 61.5 & 60.6 & 61.4 & 59.5 & 60.1 & 54.3 & 53.6 & 52.1 & 64.7 \\
        StreamTrack~\cite{StreamTrack} & 62.1 & \underline{70.8} & 38.4 & 68.6 & \underline{64.7} & \underline{66.6} & \underline{66.7} & \textbf{64.3} & \textbf{60.7} & \textbf{59.7} & 55.8 & 69.2 \\
        \hline
        ChronoTrack & \textbf{64.5} & \textbf{73.0} & \textbf{47.0} & \textbf{76.1} & \textbf{66.2} & \textbf{68.2} & \textbf{67.3} & \underline{64.0} & \underline{60.0} & \underline{57.6} & \textbf{59.7} & \textbf{72.7}  \\
        \hline
    \end{tabular}    
    }
\end{table*}
\endgroup 
\subsection{Implementation Details}

\subsubsection{Architecture.}
We adopt DGCNN~\cite{DGCNN} as the backbone, following the setups in MBPTrack~\cite{MBPTrack} and CXTrack~\cite{CXTrack}. 
It processes \(N_t = 1024\) input points through 3 EdgeConv and 3 downsampling layers, resulting in $D = 128$-dimensional features with an 8× reduction in the number of points.
We use BPLocNet as the decoder and follow its original loss formulation and training strategy from MBPTrack~\cite{MBPTrack}. 
We set the number of memory tokens to $K=32$ each with a dimension of $128$, and use $L_\text{MU}=3$ and $L_\text{MFR}=2$ layers for memory update and target-aware feature transformation, respectively.
Ablations on different \(K\) are provided in the supplementary material.

\subsubsection{Training and Inference Details.}
Following~\cite{MBPTrack}, we sample 8 consecutive frames from the point cloud sequence to form a training sample. 
The model is trained using the Adam optimizer with a learning rate of \(1 \times 10^{-3}\), and a batch size of 16. 
The learning rate is decayed by a factor of 0.2 every 15 epochs. 
We adopt bfloat16 mixed-precision training for memory efficiency.
All experiments are conducted using the PyTorch framework on two RTX 4090 GPUs.
Regarding the proposed loss terms, we set the distance threshold $\tau_{\text{dist}} = 30\,\text{cm}$ for the temporal consistency loss, and use a softmax temperature of $\tau_\text{cycle} = 0.1$ in the memory cycle consistency loss.
See the supplementary material for ablations on \(\tau_{\text{dist}}\) and \(\tau_{\text{cycle}}\).

In addition, we apply a binary mask threshold of \(\tau_{\text{mask}} = 0.5\) when filtering foreground or background points.
To stabilize the tracking under low confidence during testing, we reuse the prediction of the previous frame if the maximum predicted targetness score falls below 0.2, following MBPTrack.

\subsection{Datasets and Metrics}
\subsubsection{Datasets}
We evaluate our method on three widely used benchmarks: KITTI~\cite{KITTI}, NuScenes~\cite{NuScenes}, and Waymo Open Dataset (WOD)~\cite{Waymo}. 
KITTI provides 21 training and 29 testing sequences in 8 categories. Following~\cite{P2B}, we split the train set into sequences 0–16 for train, 17–18 for validation, and 19–20 for test, due to the absence of test annotations.
NuScenes consists of 1,000 scenes with 23 categories, captured using a 32-beam LiDAR. 
We adopt the standard 700/150/150 split for training, validation, and testing.
WOD is a large-scale dataset collected across 25 cities.
Following~\cite{STNet, MBPTrack}, we assess the generalization ability of our method on 1,121 tracklets using the KITTI pre-trained model without fine-tuning. 
The Car and Pedestrian categories in KITTI correspond to the Vehicle and Pedestrian categories in WOD.

\subsubsection{Metrics}
We adopt One Pass Evaluation~(OPE)~\cite{OPE} to assess the Success and Precision.
Success measures Area Under Curve~(AUC) of intersection over union~(IoU) between predicted and ground-truth bounding box, and Precision is the AUC of their center distance within 2 meters.
\subsection{Comparison with State-of-the-art Methods}

\paragraph{KITTI.}
In Tab.~\ref{tab:kitti}, we compare ChronoTrack with the state-of-the-art methods on the KITTI dataset.
ChronoTrack achieves the highest overall performance (Mean column).
In particular, ChronoTrack surpasses memory-based approaches such as MBPTrack, StreamTrack, M3SOT, and MemDisst by effectively incorporating long-term memory rather than relying solely on short-term context.
Moreover, while MemDisst leverages a pre-trained 2D tracker and a self-supervised 3D backbone, ChronoTrack achieves superior performance without relying on any external supervision.
In terms of efficiency, ChronoTrack runs in real-time speed of 42 FPS on a single RTX 4090 GPU while maintaining state-of-the-art performance; see the supplementary material for detailed efficiency measurements.

\begin{figure}[!t]
    \centering
    {\includegraphics[width=0.47\textwidth]{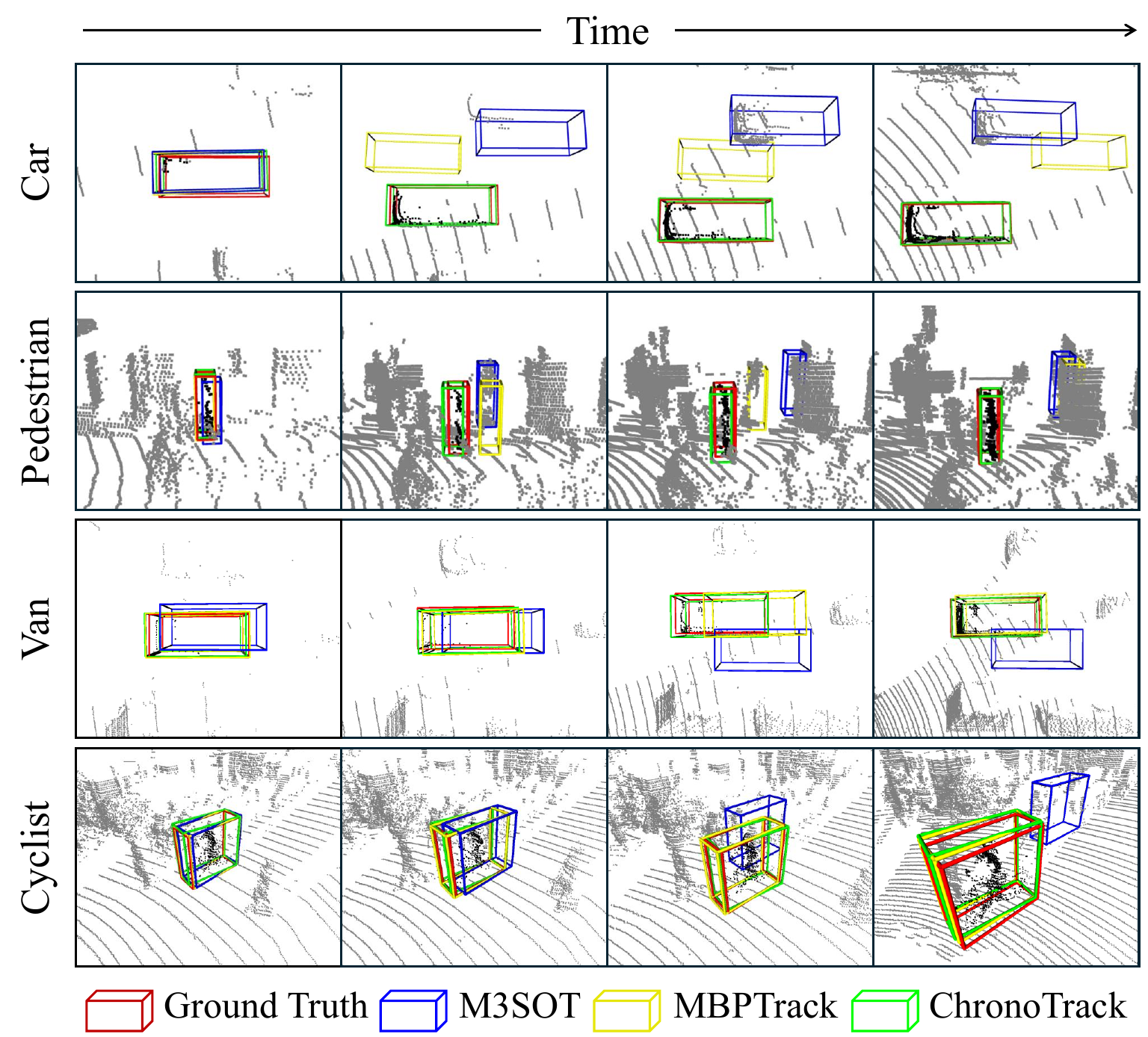}}
    \caption{
    Qualitative results on the KITTI dataset.
    }
\label{fig:fig_4}
\end{figure}

In Fig.~\ref{fig:fig_4}, we present qualitative results under challenging scenarios, including sparse observations and the presence of multiple distractors.
ChronoTrack maintains robust tracking in such cases compared to existing memory-based methods, demonstrating the efficacy of long-term memory.

\paragraph{NuScenes.}
Tab.~\ref{tab:nuScenes} shows the results on the NuScenes dataset.
ChronoTrack achieves the best overall performance on the benchmark, surpassing the second-best model, MBPTrack, by 2.2 points in Mean Success and 2.8 points in Mean Precision.
Notably, MBPTrack shares the same backbone and decoder architecture, differing only in memory design.
This highlights the effectiveness of our long-term memory in handling complex and diverse sequences.

\paragraph{WOD.}
We evaluate the generalization ability of our KITTI pre-trained model on Waymo dataset. 
Tab.~\ref{tab:waymo} shows that ChronoTrack achieves a significant performance margin over previous SOTA methods. 
This demonstrates the robustness of our method even in unseen data.

\begingroup
\setlength{\tabcolsep}{4pt} 
\renewcommand{\arraystretch}{1.0} 
\begin{table}[!t]
    \centering
    \caption{
    Comparison with the SOTA methods on WOD. The models are pre-trained on KITTI and directly tested on the WOD validation set without fine-tuning.
    }
    \label{tab:waymo}
    {
    \begin{tabular}{ l | cc | cc | cc}
        \hline
        \multirow{2}{*}{Method} &
        \multicolumn{2}{c|}{Vehicle} &
        \multicolumn{2}{c|}{Pedestrian} &
        \multicolumn{2}{c}{Mean} \\
        \cline{2-7}
        & Suc. & Prec.
        & Suc. & Prec.
        & Suc. & Prec. \\
        \hline
        P2B~\cite{P2B} & 52.6 & 35.4 & 17.9 & 29.6 & 33.0 & 33.5\\
        M$^2$-Track~\cite{M2Track} & 61.1 & 61.6 & 32.0 & 67.3 & 44.6 & 63.5 \\
        CXTrack~\cite{CXTrack} & 57.1 & 66.1 & 30.7 & 49.4 & 42.2 & 56.7 \\
        MBPTrack~\cite{MBPTrack} & 61.9 & \underline{71.9} & 33.7 & 52.7 & 46.0 & 61.0 \\
        SCVTrack~\cite{SCVTrack} & 61.3 & 69.8 & 32.2 & 50.0 & 44.8 & 58.6 \\
        M3SOT~\cite{M3SOT} & \underline{64.5} & \textbf{74.7} & 32.8 & 52.1 & \underline{46.6} &  \underline{61.9} \\
        HVTrack~\cite{HVTrack} & 59.8 & 69.7 & 30.0 & 49.1 & 43.0 & 58.1 \\
        MemDisst~\cite{MemDisst} & 61.9 & 71.8 & \underline{33.6} & \underline{52.8} & 45.9 & 61.1 \\
        \hline
        ChronoTrack & \textbf{65.5} & \textbf{74.7} & \textbf{35.2} & \textbf{55.6} & \textbf{48.4} & \textbf{63.9} \\
        \hline
    \end{tabular}
    }
\end{table}
\endgroup 

\begingroup
\setlength{\tabcolsep}{1.7pt} 
\renewcommand{\arraystretch}{1.3} 
\begin{table}[t]
    \centering
    \caption{
    Component ablation of ChronoTrack. 
    $X^{\text{FG}}$ indicates that token-level memory is used instead of point-level memory.
    $\mathcal{L}_{\text{TC}}$ and $\mathcal{L}_{\text{MCC}}$ denote the temporal consistency loss and memory cycle consistency loss, respectively.
    Success/Precision metrics are used for evaluation.
    }
    \label{tab: ablation study}
    {\footnotesize
    \begin{tabular}{ccc | c | c | c | c | c}
        \hline
        $X^{\text{FG}}$ &
        $\mathcal{L}_\text{TC}$ &
        $\mathcal{L}_\text{MCC}$ &
        Car &
        Pedestrian &
        Van &
        Cyclist &
        Mean
        \\
        \hline
        $\cdot$ & $\cdot$ & $\cdot$ & 74.0$\:\!$/$\:\!$86.2 & 67.9$\:\!$/$\:\!$93.5 & 63.6$\:\!$/$\:\!$76.1 & 77.0$\:\!$/$\:\!$94.6 & 70.5$\:\!$/$\:\!$88.6 \\
        \checkmark & $\cdot$ & $\cdot$ & 73.3$\:\!$/$\:\!$85.7 & 66.5$\:\!$/$\:\!$90.4 & 59.4$\:\!$/$\:\!$75.1 & 75.7$\:\!$/$\:\!$94.6 & 69.2$\:\!$/$\:\!$87.0 \\
        \checkmark & \checkmark & $\cdot$ & \textbf{76.1}$\:\!$/$\:\!$87.0 & 67.7$\:\!$/$\:\!$92.8 & 63.6$\:\!$/$\:\!$77.0 & 75.4$\:\!$/$\:\!$94.5 & 71.3$\:\!$/$\:\!$88.8 \\
        \checkmark & \checkmark & \checkmark  & 76.0$\:\!$/$\:\!$\textbf{88.6} & \textbf{68.6}$\:\!$/$\:\!$\textbf{94.0} & \textbf{64.2}$\:\!$/$\:\!$\textbf{77.7} & \textbf{77.7}$\:\!$/$\:\!$\textbf{94.8} & \textbf{71.8}$\:\!$/$\:\!$\textbf{90.1}  \\
        \hline
    \end{tabular}
    \vspace{-0.35cm}
    }
\end{table}
\endgroup 
\subsection{Ablation Study}
\paragraph{Component ablation.}
Tab.~\ref{tab: ablation study} presents the component-wise ablation results on the KITTI.
We begin with a baseline that adopts the short-term point-level memory used in prior works.
Applying our token-level memory to the baseline, which accumulates target features over long temporal spans, results in a performance drop.
This highlights that simply increasing the memory capacity is ineffective without addressing temporal feature inconsistency, as depicted in Fig.~\ref{fig:fig_1}.
Introducing the temporal consistency loss, which enforces alignment of target features across frames, yields a substantial improvement, demonstrating the effectiveness of temporally consistent long-term memory.
Additionally, memory cycle consistency loss further boosts performance by encouraging memory tokens to encode diverse and target-specific semantics.
This is particularly beneficial for deformable objects such as Pedestrian and Cyclist, which demand finer part-level representations to capture subtle shape variations.

\vspace{-0.2cm}
\paragraph{Ablation on various sequence lengths.}
To verify the effectiveness of our framework in long-term scenarios, we analyze performance across different temporal scales.
Tab.~\ref{tab:scale} reports the performance for four groups in the Car category on the KITTI and NuScenes datasets. 
In the case of KITTI, the sequence length ranges for each group are S~(0-15), M~(16-27), L~(28-50), and XL~(51+).
For NuScenes, the ranges are S~(0-7), M~(8-15), L~(16-26), and XL~(27+).
The consistent performance gains of ChronoTrack across all groups highlight its robustness in longer sequences, where temporal inconsistency becomes more pronounced as the token accumulates extended context.

\begingroup
\setlength{\tabcolsep}{5pt} 
\renewcommand{\arraystretch}{1.2} 
\begin{table}[t!]
    \centering
    \caption{
        Performance on various sequence lengths on KITTI and NuScenes.
        We divide all tracklets into four groups based on sequence lengths, using the 25th, 50th, and 75th percentiles as thresholds, and denote them as S, M, L, and XL. Parentheses indicate the number of tracklets per group.
    }
    \label{tab:scale}
    {\footnotesize
    \begin{tabular}{ l | cccc}
        \hline
        \multirow{2}{*}{Method} & \multicolumn{4}{c}{KITTI} \\
        & S~(29) & M~(28) & L~(28) & XL~(27) \\
        \hline
        MBPTrack~\cite{MBPTrack} & 78.0$\,$/$\,$91.5 & 73.2$\,$/$\,$85.3 & 75.4$\,$/$\,$87.6 & 72.7$\,$/$\,$83.7 \\
        ChronoTrack & \textbf{81.5}$\,$/$\,$\textbf{93.1} & \textbf{77.7}$\,$/$\,$\textbf{90.5} & \textbf{78.5}$\,$/$\,$\textbf{90.9} & \textbf{74.8}$\,$/$\,$\textbf{87.6}  \\
        Improvement & +3.5$\,$/$\,$+1.6 & +4.5$\,$/$\,$+5.2 & +3.1$\,$/$\,$+3.3 & +2.1$\,$/$\,$+3.9 \\
        \hline
        \multirow{2}{*}{Method} & \multicolumn{4}{c}{NuScenes} \\
        & S~(927) & M~(966) & L~(883) & XL~(884) \\
        \hline
        MBPTrack~\cite{MBPTrack} & 71.0$\,$/$\,$77.2 & 61.4$\,$/$\,$69.3 & 62.0$\,$/$\,$70.6 & 62.1$\,$/$\,$69.9 \\
        ChronoTrack & \textbf{72.4}$\,$/$\,$\textbf{79.3} & \textbf{63.2}$\,$/$\,$\textbf{71.5} & \textbf{62.8}$\,$/$\,$\textbf{71.6} & \textbf{65.0}$\,$/$\,$\textbf{73.5} \\
        Improvement & +1.4$\,$/$\,$+2.1 & +1.8$\,$/$\,$+2.2 & +0.8$\,$/$\,$+1.0 & +2.9$\,$/$\,$+3.6\\
        \hline
    \end{tabular}
    }
\end{table}
\endgroup 
\begin{figure}[!t]
    {\includegraphics[width=0.47\textwidth]{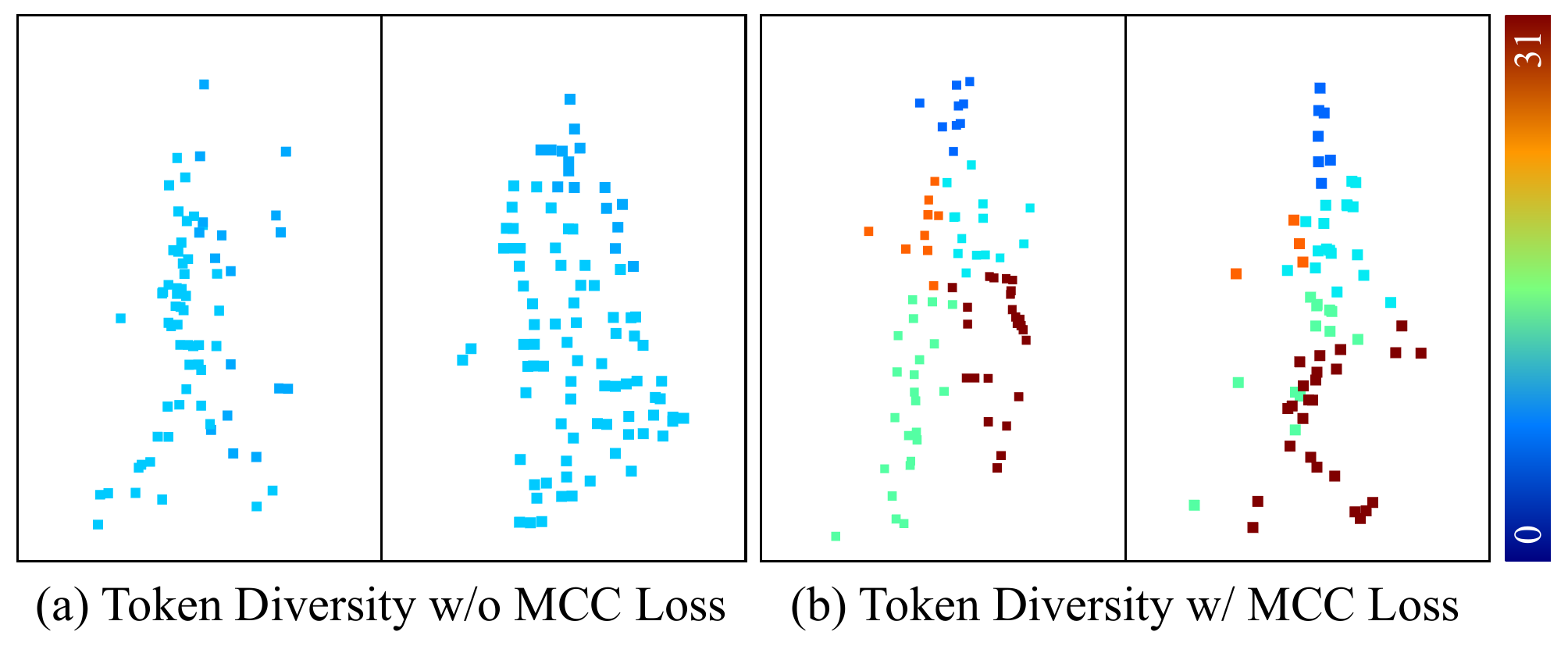}}
    \centering
    \caption{
    The diversity of the memory token.
    Color represents the index of the most similar token for each point.
    }
    \vspace{-0.2cm}
\label{fig:fig_5}
\end{figure}

\begin{figure}[t!]
    \centering
    \includegraphics[width=0.47\textwidth]{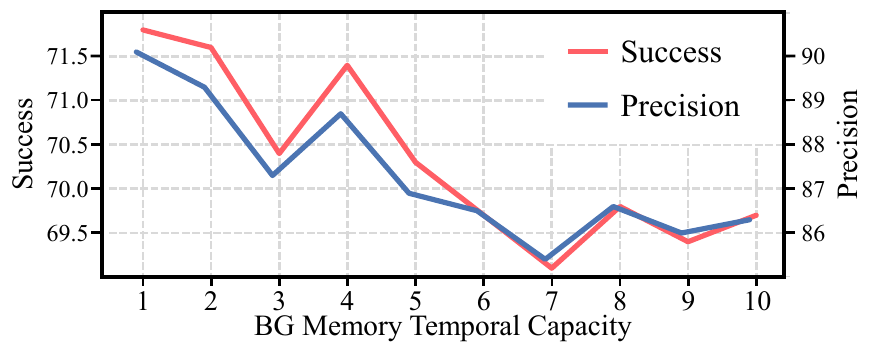}
    \caption{
        Ablation on temporal capacity of background memory.
        Mean performance on the KITTI is reported.
    }
    \label{fig:fig_6}
\end{figure}

\vspace{-0.2cm}
\paragraph{Design choice of background memory.}
To validate our short-term design of background memory, we conduct an ablation study on its temporal capacity.
While a longer context of the target in memory improves performance, as shown in Fig.~\ref{fig:fig_1}~(b), extending background memory leads to a performance decline, as shown in Fig.~\ref{fig:fig_6}.
This degradation occurs because the background that surrounds the target continuously changes, diminishing the reliability of previously accumulated background features in future frames.

\vspace{-0.2cm}
\paragraph{Diversity of memory tokens.}

In Fig.~\ref{fig:fig_5}, we visualize the diversity of memory tokens, where each token is expected to represent a distinct part of the target object.
Each point is assigned to the memory token with the highest cosine similarity, and token assignments are color-coded.
As shown in Fig.~\ref{fig:fig_5}, without the Memory Cycle Consistency~(MCC) loss, most points are assigned to a single token, whereas with the loss, various tokens are assigned to different parts of the target.
This shows that the MCC loss encourages tokens to specialize in semantically distinct regions of the target.

\begin{figure}[t!]
    \centering
    \includegraphics[width=0.47\textwidth]{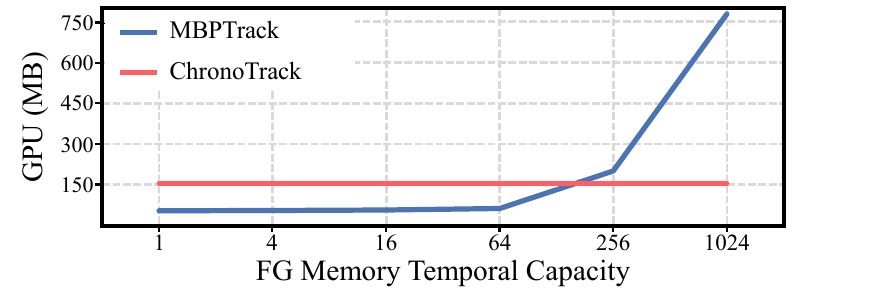}
    \caption{
        Comparison of GPU memory overhead as the temporal capacity of foreground memory increases.
    }
    \vspace{-0.3cm}
    \label{fig:fig_7}
\end{figure}

\vspace{-0.2cm}
\paragraph{Scalability.}
Fig.~\ref{fig:fig_7} compares the GPU memory consumption of MBPTrack~\cite{MBPTrack} and ChronoTrack as the temporal capacity of memory increases.
While MBPTrack exhibits continuous growth in GPU memory usage with larger temporal capacity, ChronoTrack achieves superior scalability by maintaining a constant memory footprint.

\section{Conclusion}
In this paper, we identified that existing memory-based 3D-SOT methods struggle to scale beyond short-term context due to temporal inconsistency and inefficient memory structures.  
To overcome this, we proposed \textit{ChronoTrack}, a robust 3D-SOT framework that employs token-level long-term memory with two complementary losses to enforce temporal consistency and promote token diversity.  
Extensive experiments demonstrate that ChronoTrack effectively leverages long-range temporal cues and achieves state-of-the-art performance across multiple benchmarks.

\section*{Acknowledgements}
This work was supported in part by MSIT/IITP (No. RS-2022-II220680, RS-2020-II201821, RS-2019-II190421, RS-2024-00459618, RS-2024-00360227, RS-2024-00437633, RS-2024-00437102, RS-2025-25442569), MSIT/NRF (No. RS-2024-00357729), and KNPA/KIPoT (No. RS-2025-25393280).

{
    \small
    \bibliographystyle{ieeenat_fullname}
    \bibliography{main}
}

\clearpage
\setcounter{page}{1}
\maketitlesupplementary

\section{Hyperparamter Choices}
\label{sec:supple}

\begingroup
\setlength{\tabcolsep}{3.6pt} 
\renewcommand{\arraystretch}{1.3} 
\begin{table}[t]
    \centering
    \caption{
    Performance comparison with different numbers of memory tokens $K$ on the KITTI dataset.
    }
    \label{tab: number of token ablation}
    {\footnotesize
    \begin{tabular}{c | c | c | c | c | c}
        \hline
        Setting & Car & Pedestrian & Van & Cyclist & Mean \\
        \hline
        $K=16$ & 74.8$\:\!$/$\:\!$86.8 & 68.6$\:\!$/$\:\!$93.0 & 62.4$\:\!$/$\:\!$76.7 & \textbf{78.1}$\:\!$/$\:\!$\textbf{94.8} & 71.1$\:\!$/$\:\!$88.7 \\        
        $K=32$ & \textbf{76.0}$\:\!$/$\:\!$\textbf{88.6} & \textbf{68.6}$\:\!$/$\:\!$\textbf{94.0} & \textbf{64.2}$\:\!$/$\:\!$\textbf{77.7} & 77.7$\:\!$/$\:\!$\textbf{94.8} & \textbf{71.8}$\:\!$/$\:\!$\textbf{90.1} \\
        $K=64$ & 74.8$\:\!$/$\:\!$86.7 & 67.7$\:\!$/$\:\!$92.8 & 63.5$\:\!$/$\:\!$76.4 & 76.7$\:\!$/$\:\!$94.4 & 70.8$\:\!$/$\:\!$88.6 \\
        \hline
    \end{tabular}
    }
\end{table}
\endgroup 
\begingroup
\setlength{\tabcolsep}{3pt} 
\renewcommand{\arraystretch}{1.3} 
\begin{table}[t]
    \centering
    \caption{
    Performance comparison with different temperature $\tau_{\text{cycle}}$ of memory cycle consistency loss on the KITTI dataset.
    }
    \label{tab: temperature ablation}
    {\footnotesize
    \begin{tabular}{@{}r@{$\;\!=$$\;\!$}l | c | c | c | c | c}
    \hline
    \multicolumn{2}{c|}{Setting} & Car & Pedestrian & Van & Cyclist & Mean \\
    \hline
    $\tau_{\text{cycle}}$ & 0.07 & 74.9$\:\!$/$\:\!$86.9 & \textbf{68.7}$\:\!$/$\:\!$93.2 & 59.7$\:\!$/$\:\!$74.4 & 76.2$\:\!$/$\:\!$94.7 & 70.9$\:\!$/$\:\!$88.7 \\
    $\tau_{\text{cycle}}$ & 0.1  & \textbf{76.0}$\:\!$/$\:\!$\textbf{88.6} & 68.6$\:\!$/$\:\!$\textbf{94.0} & \textbf{64.2}$\:\!$/$\:\!$77.7 & \textbf{77.7}$\:\!$/$\:\!$\textbf{94.8} & \textbf{71.8}$\:\!$/$\:\!$\textbf{90.1} \\
    $\tau_{\text{cycle}}$ & 1.0  & 73.6$\:\!$/$\:\!$86.2 & 68.2$\:\!$/$\:\!$92.4 & 63.9$\:\!$/$\:\!$\textbf{78.2} & 76.5$\:\!$/$\:\!$\textbf{94.8} & 70.4$\:\!$/$\:\!$88.4 \\
    \hline
    \end{tabular}
    }
\end{table}
\endgroup 
\begingroup
\setlength{\tabcolsep}{2.7pt} 
\renewcommand{\arraystretch}{1.3} 
\begin{table}[t]
    \centering
    \caption{
    Performance comparison with different distance thresholds $\tau_{\text{dist}}$ on the KITTI dataset.
    }
    \label{tab: distance threshold}
    {\footnotesize
    \begin{tabular}{c | c | c | c | c | c}
        \hline
        Setting & Car & Pedestrian & Van & Cyclist & Mean \\
        \hline
        $\tau_{\text{dist}}=10cm$ & 74.9$\:\!$/$\:\!$86.7 & \textbf{69.4}$\:\!$/$\:\!$93.3 & 64.1$\:\!$/$\:\!$\textbf{78.0} & 76.6$\:\!$/$\:\!$94.6 & 71.6$\:\!$/$\:\!$89.0 \\        
        $\tau_{\text{dist}}=30cm$ & \textbf{76.0}$\:\!$/$\:\!$\textbf{88.6} & 68.6$\:\!$/$\:\!$\textbf{94.0} & \textbf{64.2}$\:\!$/$\:\!$77.7 & \textbf{77.7}$\:\!$/$\:\!$\textbf{94.8} & \textbf{71.8}$\:\!$/$\:\!$\textbf{90.1} \\
        $\tau_{\text{dist}}=70cm$ & 74.5$\:\!$/$\:\!$86.8 & 67.9$\:\!$/$\:\!$92.9 & 63.8$\:\!$/$\:\!$77.6 & 76.9$\:\!$/$\:\!$\textbf{94.8} & 70.8$\:\!$/$\:\!$88.8 \\
        \hline
    \end{tabular}
    }
\end{table}
\endgroup 

We conduct ablation studies on different hyperparameter choices: number of memory tokens $K$, temperature of memory cycle consistency loss $\tau_{\text{cycle}}$, and distance threshold  of temporal consistency loss $\tau_{\text{dist}}$.
In ~\cref{tab: number of token ablation}, we find that $K=32$ is suitable for overall performance.
~\cref{tab: temperature ablation} shows performances for the different temperatures of the memory cycle consistency loss.
The temperature $\tau_{\text{cycle}}$ controls the sharpness of the transition probability distribution during the cyclic walk.
As shown, $\tau_{\text{cycle}} = 0.1$ yields the best performance.
~\cref{tab: distance threshold} varies the distance threshold \(\tau_{\text{dist}}\) in the temporal consistency loss. 
\(\tau_{\text{dist}}\) is used to discard nearest neighbor pairs whose canonical distance exceeds the threshold to suppress unreliable matches.
We find \(\tau_{\text{dist}} = 30\,\text{cm}\) provides the best overall performance.

\begingroup
\setlength{\tabcolsep}{2.7pt} 
\renewcommand{\arraystretch}{1.3} 
\begin{table}[t]
    \centering
    \caption{
    Comparison of model size and inference time with state-of-the-art methods.
    }
    \label{tab: complexity}
    {\footnotesize
    \begin{tabular}{l | c | c | c | c}
        \hline
        Mehod & \# Params & Inference Time & FPS & Suc./Prec. \\
        \hline
        MBPTrack~\cite{MBPTrack} & 1.9M & 15ms & 67 & 73.4$\:\!$/$\:\!$84.8 \\ 
        M3SOT~\cite{M3SOT} & 4.3M & 48ms & 21 & 75.9$\:\!$/$\:\!$87.4 \\
        ChronoTrack & 2.9M & 24ms & 42 & 76.0$\:\!$/$\:\!$88.6\\
        \hline
    \end{tabular}
    }
\end{table}
\endgroup 

\section{Model Size and Inference Time}
\label{sec:complexity}
~\cref{tab: complexity} compares model size and runtime on the KITTI Car category, evaluated on a single RTX 4090 GPU using official implementations of the compared state of the art methods.
ChronoTrack processes a frame in 24 ms (42 FPS) with 2.9M parameters, achieving real time performance and the highest Success/Precision among the compared methods (76.0/88.6).

\end{document}